\documentclass[letterpaper]{article} 
\usepackage{aaai2026}
\usepackage{times}  
\usepackage{helvet}  
\usepackage{courier}  
\usepackage[hyphens]{url}  
\usepackage{graphicx} 
\usepackage{natbib}  
\usepackage{caption} 
\usepackage{algorithm}
\usepackage{algorithmicx}
\usepackage{algpseudocode}
\usepackage{amsmath}
\usepackage{amsfonts}
\usepackage{booktabs}
\usepackage{newfloat}
\usepackage{listings}
\DeclareCaptionStyle{ruled}{labelfont=normalfont,labelsep=colon,strut=off} 
\floatstyle{ruled}
\newfloat{listing}{tb}{lst}{}
\floatname{listing}{Listing}
\title{PREF: Reference‑Free Evaluation of Personalised Text Generation in LLMs}
\author {
   Xiao Fu\textsuperscript{\rm 1},
   Hossein A. Rahmani\textsuperscript{\rm 1},
   Bin Wu\textsuperscript{\rm 1},
   Jerome Ramos\textsuperscript{\rm 1},
   Emine Yilmaz\textsuperscript{\rm 1},
   Aldo Lipani\textsuperscript{\rm 1}
}
\affiliations {
    \textsuperscript{\rm 1}University College London (UCL)\\
    xiao.fu.20@ucl.ac.uk, hossein.rahmani.22@ucl.ac.uk, bin.wu.23@ucl.ac.uk, jerome.ramos.20@ucl.ac.uk, emine.yilmaz@ucl.ac.uk, aldo.lipani@ucl.ac.uk
}

\usepackage{bibentry}

\begin{document}

\maketitle

\begin{abstract}

Personalised text generation is essential for user-centric information systems, yet most evaluation methods overlook the individuality of users. We introduce \textbf{PREF}, a \textbf{P}ersonalised \textbf{R}eference-free \textbf{E}valuation \textbf{F}ramework that jointly measures general output quality and user-specific alignment without requiring gold personalised references. PREF operates in a three-step pipeline: (1) a \emph{coverage stage} uses a large language model (LLM) to generate a comprehensive, query-specific guideline covering universal criteria such as factuality, coherence, and completeness; (2) a \emph{preference stage} re-ranks and selectively augments these factors using the target user’s profile, stated or inferred preferences, and context, producing a personalised evaluation rubric; and (3) a \emph{scoring stage} applies an LLM judge to rate candidate answers against this rubric, ensuring baseline adequacy while capturing subjective priorities. This separation of coverage from preference improves robustness, transparency, and reusability, and allows smaller models to approximate the personalised quality of larger ones. Experiments on the PrefEval benchmark, including implicit preference-following tasks, show that PREF achieves higher accuracy, better calibration, and closer alignment with human judgments than strong baselines. By enabling scalable, interpretable, and user-aligned evaluation, PREF lays the groundwork for more reliable assessment and development of personalised language generation systems.
\end{abstract}

\section{Introduction} \label{sec:intro}

Large language models (LLMs) such as GPT-3~\cite{brown2020gpt3} and ChatGPT~\cite{openai2023chatgpt} have propelled open-ended text generation to new heights, enabling high-quality dialogue, code synthesis, and data-to-text narration at scale.  Despite these successes, \emph{evaluating} the outputs of such models remains an open problem.  Traditional reference-based metrics, including BLEU and ROUGE~\cite{papineni2002bleu,lin2004rouge}, which count $n$-gram overlap with one or more gold references, correlate weakly with human judgements on tasks that admit many equally valid answers (e.g., creative writing, recommendation, and advice).  Embedding-based alternatives such as BERTScore~\cite{zhang2019bertscore} mitigate surface mismatch by measuring semantic similarity in a latent space, yet they still require high-quality reference texts and ultimately reflect \emph{generic} rather than \emph{user-specific} desiderata.

A parallel line of research bypasses references altogether by treating a strong LLM as an automatic judge.  Benchmarks such as MT-Bench~\cite{zheng2023judging} and AlpacaEval~\cite{li2023alpacaeval} ask GPT-4 to rate or rank candidate answers and report impressive agreement with crowd workers.  Although this \emph{LLM-as-a-judge} paradigm scales cheaply and reproducibly, its rubric is fundamentally universal---``a good answer is a good answer for everyone''---and therefore blind to individual preferences, constraints, or prior interactions that might radically alter a user’s perception of quality.

Modern applications increasingly expose LLMs directly to end users who expect outputs tailored to their tastes, goals, and contexts.  Recent work has shown that LLMs can indeed adapt to user profiles, feedback signals, and stylistic norms~\cite{zhang2024personalization}. However, evaluating such \emph{personalised} generation is especially tricky because quality is now user-dependent, i.e., \textit{an answer that delights one person might frustrate another}. Generic references or model-centric rubrics cannot capture this subjective dimension.

Human evaluation remains the gold standard but is expensive, time-consuming, and prone to annotator variance and bias---even when detailed guidelines are provided~\cite{vanderlee2019best}. Collecting user-aligned ratings for every system update or user cohort is therefore impractical, hindering rapid iteration on personalised experiences.

To fill this evaluation gap, we propose \textbf{PREF}, a \textbf{P}ersonalised, \textbf{R}eference-free \textbf{E}valuation \textbf{F}ramework that scores generated text \emph{without} gold personalised answers. \textsc{PREF} evaluates in two stages:

\begin{enumerate}
    \item \textbf{General-quality stage.}  A general guideline enumerates the salient factors to check---truthfulness, coherence, completeness, etc.---\emph{explicitly ignoring} user preferences to ensure baseline adequacy.
    \item \textbf{User-alignment stage.}  A personalised guideline is synthesised from user information (profile attributes, past dialogue, stated preferences) and used to weight or re-rank the general factors, yielding a customised rubric that reflects what \emph{this} user cares about.
\end{enumerate}

The LLM then judges candidate answers against the composite rubric, producing a scalar score without ever consulting ground-truth references. In this way, PREF combines universal quality control with user-centric alignment while remaining scalable and reproducible.

Our contribution can be summarised as:
\begin{itemize}
    \item We formulate a \emph{reference-free} evaluation protocol for personalised generation that jointly considers task adequacy and user alignment.
    \item We instantiate this protocol with automatic rubric construction and LLM-based scoring, eliminating the need for gold references or repeated human ratings.
    \item Through extensive experiments on the PrefEval benchmark, we demonstrate that PREF's scores track human judgements of personalised quality more faithfully than existing baselines.
    \item We show that integrating PREF during development helps smaller models (e.g., LLaMA-3 8B) close much of the performance gap to larger counterparts, facilitating cost-effective deployment.
\end{itemize}

Taken together, our findings position PREF as a practical and effective foundation for evaluating---and ultimately improving---user-aligned language generation systems.

\section{Related Work}
\label{sec:related}

\subsection{Foundations of Personalisation}
The idea of personalisation comes from two related areas of research:

\paragraph{Explicit user modelling.}
Adaptive hypermedia systems demonstrated that maintaining a \emph{fine-grained} learner or reader model enables real-time adaptation of content, sequencing, and navigation~\cite{Brusilovsky2001Adaptive}.  These systems relied on explicitly encoded attributes—background knowledge, goals, learning style—to decide what to show next.

\paragraph{Implicit preference inference.}
Collaborative filtering techniques such as \textsc{GroupLens} showed that preferences can be inferred from \emph{behavioural signals}: user–item ratings aggregated at scale allow “people who agreed in the past to agree again”~\cite{Resnick1994GroupLens}.  

Modern LLM personalisation strategies still trace back to this dichotomy: \emph{explicit} profile conditioning versus \emph{implicit} signal mining (e.g.\ clicks, dwell time, edits).

\subsection{Personalised Text Generation with LLMs}

\paragraph{Early neural persona modelling.}
The \textsc{Persona-Chat} framework appended vector-based speaker embeddings to seq-to-seq dialogue models, boosting persona consistency and opening the door to persona-aware generation~\cite{li2016persona}.

\paragraph{Attribute and style control.}
Plug-and-Play Language Models (PPLM) pioneered gradient-based steering of frozen GPT-2 towards user-specified attributes without expensive full fine-tuning~\cite{dathathri2020pplm}.  Subsequent \emph{parameter-efficient} methods such as prefix-tuning, adapters, and LoRA attach small trainable modules, enabled per-user or per-task specialisation with a few thousand parameters~\cite{li2021prefixtuning}.

\paragraph{Personalised RLHF and PEFT.}
Reinforcement Learning from Human Feedback (RLHF) has recently been adapted to heterogeneous user groups: \textit{Personalised-RLHF} trains multiple reward heads or preference models, each aligned with a user cluster~\cite{li2024prlhf}.  In parallel, lightweight PEFT modules distribute personalisation across thousands of users at commodity cost~\cite{tan2024democratizing}.

\paragraph{Prompt- and profile-based conditioning.}
Instead of opaque embeddings, several works propose \emph{editable natural-language profiles} that the model can quote, justify, or critique, improving scrutability in recommendation and dialogue~\cite{ramos2024nlprofiles}.  Soft-prompt methods such as \textsc{PeaPOD} blend collaborative and individual signals into one prompt, achieving strong personalisation with no model updates~\cite{ramos2024peapod}.

\subsection{Evaluating Personalisation in LLMs}

\paragraph{Limitations of reference metrics.}
Reference-based scores (BLEU, ROUGE) capture lexical overlap, and even semantic metrics like BERTScore remain insensitive to \emph{who} the answer is for.  Consequently, they correlate poorly with user-perceived quality in personalised settings.

\paragraph{LLM-as-judge approaches.}
Automatic judges such as AuPEL use a strong LLM to score \emph{personalisation}, \emph{relevance}, and \emph{fluency} jointly, removing the need for references~\cite{wang2023aupel}.  PerSE further conditions the evaluator on an explicit preference profile, increasing correlation with human ratings on open-ended tasks~\cite{wang2024perse}.

\paragraph{Benchmark suites.}
To stress-test models, several benchmarks target specific facets:

\begin{itemize}
    \item \textbf{LaMP} and \textbf{LongLaMP} evaluate short-form and long-form personalised generation, respectively (classification, email, reviews)~\cite{salemi2023lamp,kumar2024longlamp}.
    \item \textbf{PersonaLens} simulates rich multi-session profiles and scores task success, personalisation, and quality via paired LLM agents~\cite{Zhao2025PersonaLens}.
    \item \textbf{PersonaMem} tracks performance as user preferences \emph{evolve}, revealing that most LLMs struggle with drifting personas~\cite{Jiang2025PersonaMem}.
    \item \textbf{PersoBench} focuses on persona-driven chit-chat, reporting that models often ignore or contradict supplied personas even with chain-of-thought prompting~\cite{Afzoon2024PersoBench}.
    \item \textbf{PrefEval} probes adherence to both \emph{explicit} and \emph{implicit} preferences in conversational QA; accuracy degrades sharply as context length grows~\cite{Zhao2025PrefEval}.
\end{itemize}

\subsection{Positioning of This Work}
Despite rapid progress, existing evaluators either (i) demand labour-intensive human annotations, (ii) treat all users identically, or (iii) measure only narrow aspects of quality.  \textbf{PREF} addresses these gaps by:

\begin{enumerate}
    \item Constructing a \emph{two-tier} rubric that fuses universal quality checks with user-specific priorities,
    \item Operating \emph{reference-free}, ensuring scalability and reproducibility, and
    \item Remaining model-agnostic, thereby supporting diverse backbone LLMs and personalisation mechanisms.
\end{enumerate}

In the experiments that follow, we concentrate on open-domain question answering and adopt the \textbf{PrefEval} benchmark as our primary testbed, as it offers both explicit and implicit preference challenges that align with PREF’s design goals.




\section{From Personalisation to Evaluation} \label{sec:pref}
\begin{algorithm}
\caption{PREF Scoring Pipeline}
\label{algo:pref_pipeline}
\begin{algorithmic}
\Require Question set $Q=\{q_n\}_{n=1}^N$, \\
Preference set $P=\{p_n\}_{n=1}^N$, \\
Answer set $A=\{a_n\}_{n=1}^N$, \\
Coverage LLM $\mathcal{M}_{\text{cov}}$, \\
Preference LLM $\mathcal{M}_{\text{pref}}$, \\
Scoring LLM $\mathcal{M}_{\text{score}}$
\Ensure Personalized scores $S=\{s_n\}_{n=1}^N$

\State \textbf{Phase 1: Generate General Guidelines}
\For{each $q_n \in Q$}
    \State $g_n \gets \textsc{GenerateGuideline}(\mathcal{M}_{\text{cov}}, q_n)$ \\\Comment{$g_n$ lists factors $\{f_{n,1},\dots,f_{n,K_n}\}$}
\EndFor

\State \textbf{Phase 2: Personalize Guidelines}
\For{each index $n$ from $1$ to $N$}
    \State $g^{\!*}_n \gets \textsc{Personalize}(\mathcal{M}_{\text{pref}}, q_n, p_n, g_n)$ \\\Comment{Re-rank / augment factors to obtain personalized guideline}
\EndFor
\State \textbf{Phase 3: Score Answers}
\For{each index $n$ from $1$ to $N$}
    \State $s_n \gets \textsc{ScoreAnswer}(\mathcal{M}_{\text{score}}, q_n, p_n, g^{\!*}_n, a_n)$
\EndFor
\State \Return $S$
\end{algorithmic}
\end{algorithm}
\begin{figure*}[tbp]
    \centering
    \includegraphics[width=\linewidth]{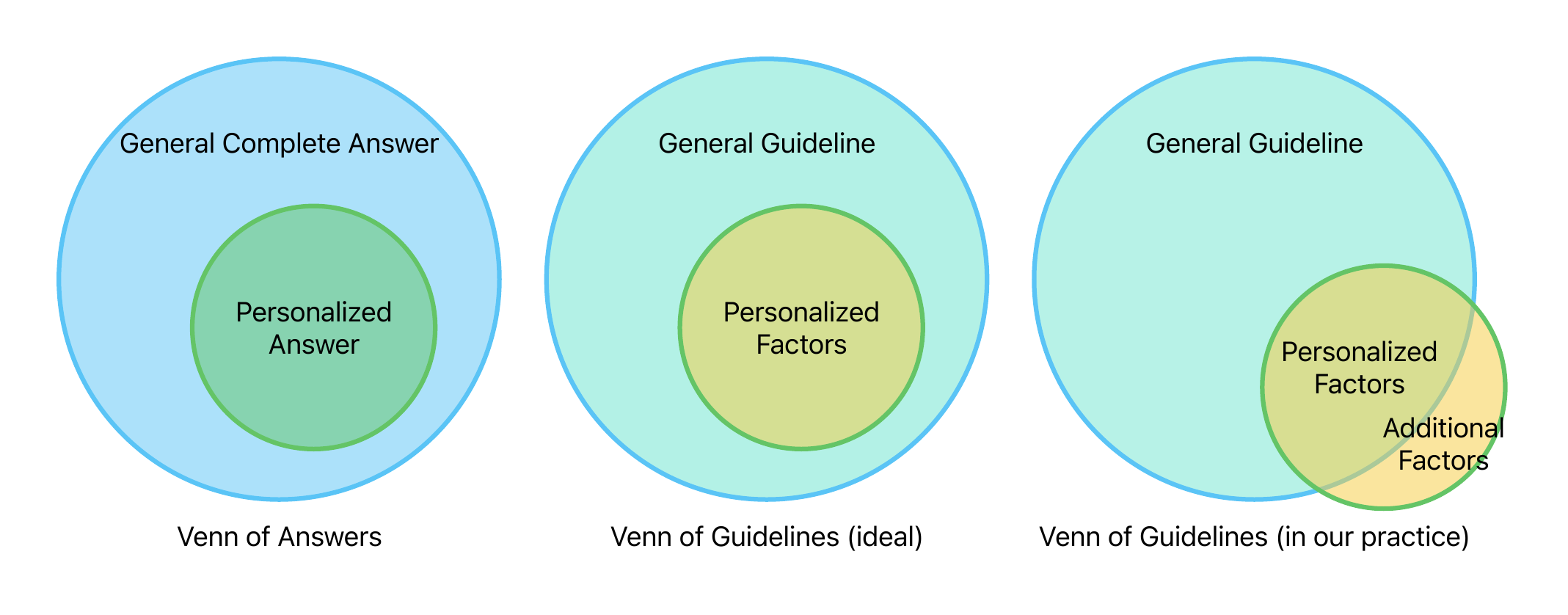}
    \caption{\textbf{\textsc{PREF} at a glance.} Venn diagrams illustrating (left) the relationship between an \emph{ideal, exhaustive} answer and a \emph{personalised} answer to the same query; (centre) the relationship between an \emph{ideal, general} guideline and a \emph{user-specific} selection (or ordering) of factors; and (right) the practical regime adopted by \textsc{PREF}, where the ranker may add extra factors to craft a personalised guideline when the general one is incomplete.}
    \label{fig:venn_personalised}
\end{figure*}

\begin{figure*}[ht]
    \centering
    \includegraphics[width=\linewidth]{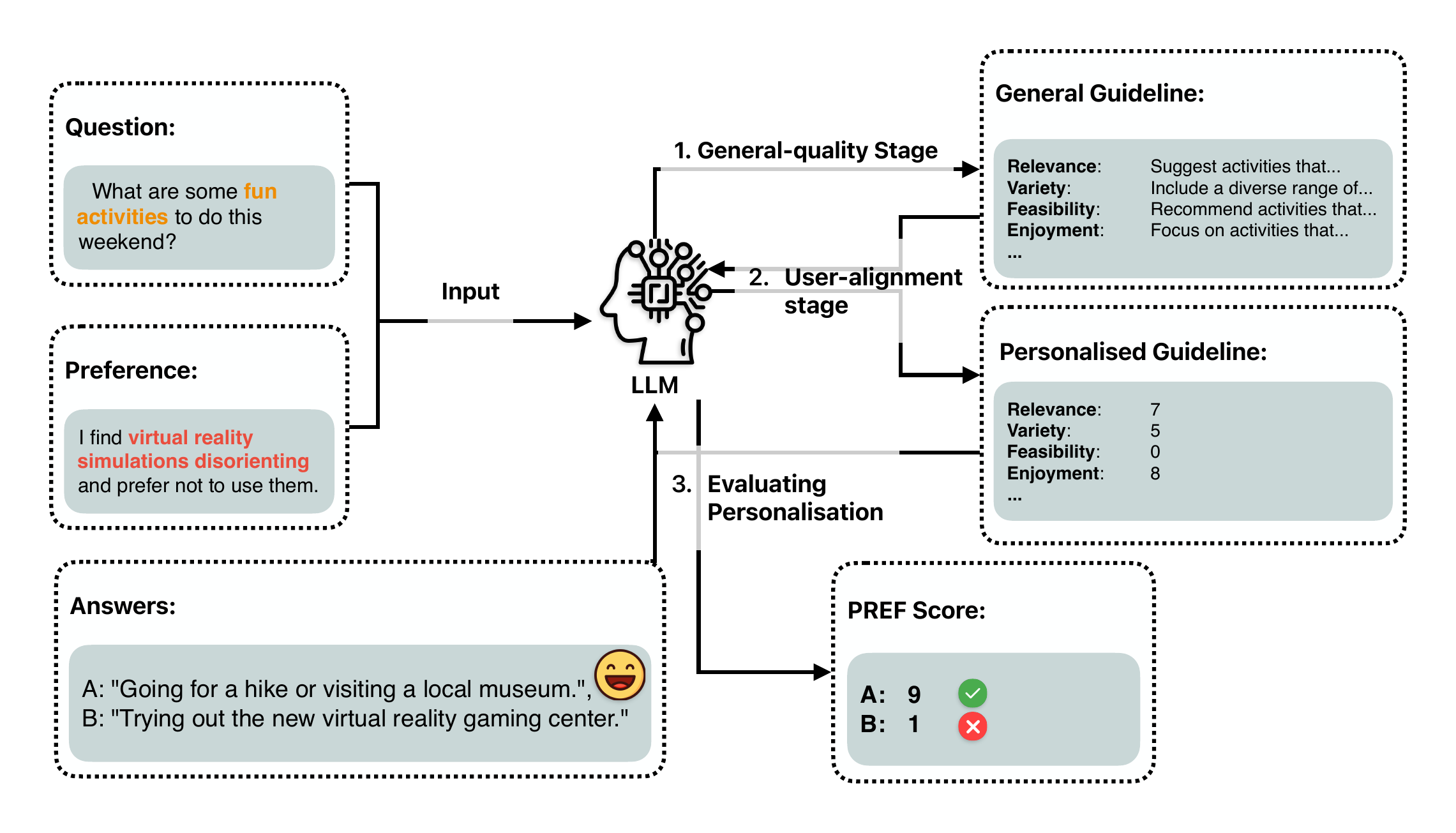}
    \caption{\textbf{The \textsc{PREF} scoring pipeline.} A three-step process—coverage, preference (User profiles), and scoring—maps a query \(q\), user profile \(p\), and candidate answer \(a\) to a final personalised score \(s_a\) with a real example on PrefEval.}
    \label{fig:pipeline}
\end{figure*}

\emph{Personalisation}—“the act of adapting something to suit the needs of a particular person”\footnote{\url{https://dictionary.cambridge.org/dictionary/english/personalization}}—is most valuable when it operates within the affordances of the underlying system. \citet{kirk2024benefits} articulate this frontier: inside a technology’s general capabilities, personalisation occupies the zone where users can adapt the system to their values, preferences, or information needs.  In other words, a system may support many behaviours in principle, but only a subset will be \emph{useful} for a given individual in a given context.

\paragraph{Design principles behind \textsc{PREF}.}
Building on this view, \textsc{PREF} rests on two assumptions that make the design space precise:
\begin{itemize}
    \item \textbf{Assumption~1 (Unlimited attention).} If users had unlimited attention and patience, a \emph{single}, fully comprehensive answer would satisfy any query, making explicit personalisation unnecessary.
    \item \textbf{Assumption~2 (Bounded attention).} In reality, attention and patience are scarce. Users therefore prefer answers that \emph{foreground} information aligned with their personal preferences, even if this omits less relevant details.
\end{itemize}

Figure~\ref{fig:venn_personalised} visualises these assumptions. The left panel contrasts an exhaustive answer with a lean, prioritised one; the centre panel contrasts an all-purpose guideline with the subset (or ranking) that matters to a particular user; and the right panel shows how, in practice, the ranker may \emph{inject} additional factors when the general guideline overlooks something salient.  This last permission reflects a pragmatic reality: current LLMs can miss important criteria during coverage; allowing augmentation prevents these omissions from propagating into the final score.

\paragraph{From principles to practice.}
\textsc{PREF} translates the above into a modular pipeline with clear interfaces.  We denote by \(g\) a \emph{general} guideline for query \(q\) that enumerates salient evaluation factors \(\mathcal{F}=\{f_1,\dots,f_n\}\).  A personalised guideline \(g^{\!*}\) is then obtained by re-ordering or re-weighting \(\mathcal{F}\) using information from a user profile \(p\).  Finally, a scoring component evaluates a candidate answer \(a\) against \(g^{\!*}\) in the context of \((q,p)\) to produce a scalar \(s_a \in \mathbb{R}\).  Intuitively, \(g\) ensures \emph{coverage} (we know what to check) and \(g^{\!*}\) ensures \emph{alignment} (we know what to check \emph{first} and how much it matters).

\paragraph{Two-stage evaluation framework.}
\textsc{PREF} operationalises this idea in two stages, followed by scoring (Figure~\ref{fig:pipeline}):
\begin{enumerate}
    \item \textbf{Coverage stage (\(q \!\to\! g\)).} Given a query \(q\), a \emph{coverage LLM} produces a comprehensive guideline \(g\) that enumerates salient factors \(\mathcal{F}\).  Each factor can optionally include a short rubric or test, making \(g\) directly actionable.\footnote{Figure~\ref{fig:pipeline} shows an example of \(g\).}
    \item \textbf{Preference stage (\(p \!\to\! \pi\) or \(\mathbf{w}\)).} A \emph{preference LLM}, conditioned on user profile \(p\), induces an ordering \(\pi\) over \(\mathcal{F}\) or a weight vector \(\mathbf{w}\in\mathbb{R}^{|\mathcal{F}|}_{\ge0}\).  When the coverage is incomplete, the ranker may \emph{augment} \(\mathcal{F}\) with additional factors derived from \(p\) or the query context, yielding \(g^{\!*}\).
    \item \textbf{Scoring stage (\(a \!\mapsto\! s_a\)).} A \emph{scoring LLM} assesses \(a\) against \(g^{\!*}\) in the context of \((q,p)\), producing a final personalised score \(s_a\).  We write \(s_a = \mathrm{Score}(a \mid q, p, g^{\!*})\).
\end{enumerate}

\paragraph{Why two stages?}
Separating \emph{coverage} from \emph{preference} yields three practical advantages:
\begin{enumerate}
    \item \textbf{Robustness to omission.} Coverage is query-specific and model-agnostic; the preference stage can add missing but user-critical factors.
    \item \textbf{Transparency and control.} The factors in \(g\) (and their user-conditioned ordering in \(g^{\!*}\)) form a human-auditable rubric that explains \emph{why} a score was assigned.
    \item \textbf{Reusability.} A single \(g\) can support many users (varying \(p\)), and a single \(p\) can be applied across related queries (varying \(q\)), reducing recomputation and enabling ablations.
\end{enumerate}

\paragraph{Scoring behaviour and constraints.}
In practice, the scorer balances universal desiderata (e.g., correctness, completeness, clarity) with user-specific ones (e.g., dietary constraints, tone, budget).  The design enforces two invariants: (i) user alignment must never \emph{excuse} factual errors introduced by the model, and (ii) augmentation may not contradict the coverage factors without justification.  These guardrails help maintain reliability while still capturing individual preferences.

Figure~\ref{fig:pipeline} and Algorithm~\ref{algo:pref_pipeline} detail the flow. Given \(q\), \(p\), and a candidate answer \(a\):
\begin{enumerate}
    \item A \emph{coverage LLM} generates \(g\).
    \item A \emph{preference LLM} re-ranks (and, if necessary, augments) \(g\) to obtain \(g^{\!*}\).
    \item A \emph{scoring LLM} assesses \(a\) against \(g^{\!*}\) in the context of \((q,p)\), producing the personalised score \(s_a\).
\end{enumerate}
To simplify, we assign one backbone LLM to take all three roles.

\section{Experiments and Findings}
\label{sec:exp}


\subsection{Data}
To assess the performance of \textsc{PREF}, we adopt the \textbf{PrefEval} benchmark, which supplies \emph{question–preference–answer} triples tailored for personalised text generation~\cite{Zhao2025PrefEval}.  We focus on the \emph{implicit multiple‐choice} subset of PrefEval, where the link between user preferences and the “ideal” answer is not stated explicitly.  For instance, if a user dislikes fish, a question asking for a recommended Japanese dish is implicitly challenging because many Japanese dishes contain fish even though the question never mentions it.  Each example offers four candidate answers, only one of which a human annotator marks as satisfactory.  

Following common practice, we hold out 20\,\% of the data as a test set and report all metrics on this split, including 200 questions, 200 user preferences and 800 answers.

\subsection{LLMs}
To probe the flexibility of \textsc{PREF}, we pair it with four different backbone language models:

\begin{itemize}
    \item \textbf{Claude 3 Haiku: claude-3-haiku-20240307} ,
    \item \textbf{GPT-4.1 Mini: gpt-4.1-mini},
    \item \textbf{LLaMA 3 8B: llama3-8b-instruct}, and
    \item \textbf{LLaMA 3 70B: llama3-70b-instruct}.
\end{itemize}

For every model, we set the sampling temperature to~0 to guarantee deterministic generation and full reproducibility.

\subsection{Evaluating Personalisation}
\label{sec:exp1}

\begin{figure*}[t]
    \centering
    \includegraphics[width=\linewidth]{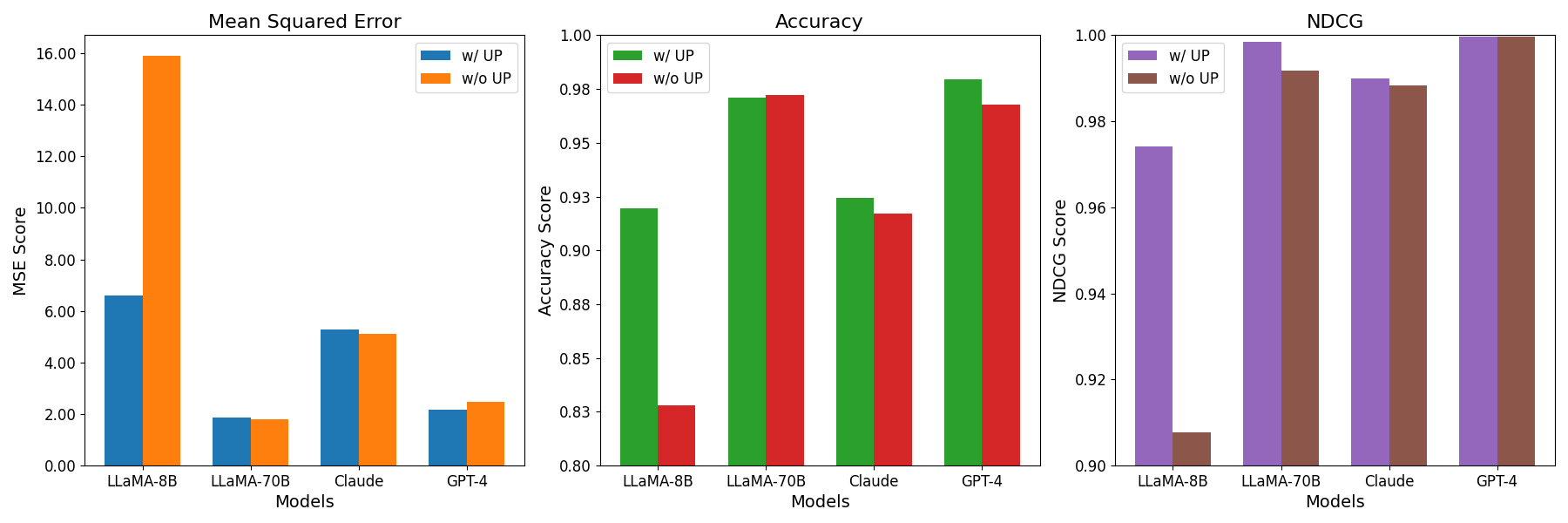}
    \caption{\textbf{Ablation on the user-preference (UP).}  \textbf{w/ UP}: full \textsc{PREF} pipeline; \textbf{w/o UP}: answers are scored only against the \emph{general} guideline from Stage one.  Bars report accuracy, MSE (lower is better), and nDCG (higher is better) on the \textit{PrefEval} implicit multiple-choice split.}
    \label{fig:w_wo_up}
\end{figure*}

\begin{table}[t]
    \centering
    
    \setlength{\tabcolsep}{6pt} 
    \resizebox{\columnwidth}{!}{
    \begin{tabular}{lccc}
        \toprule
        \textbf{Method} & \textbf{LLaMA\,3 8B} & \textbf{LLaMA\,3 70B} & \textbf{Claude 3 Haiku} \\
        \midrule
        Zero-shot        & 27 & 37 & 34 \\
        Reminder         & 84 & 91 & 88 \\
        \textsc{PREF}\,(ours) & \textbf{92} & \textbf{97} & \textbf{92} \\
        \bottomrule
    \end{tabular}
    }
    \caption{Accuracy (\%) on the PrefEval \emph{implicit multiple-choice} subset. Bold numbers denote the best score for each backbone model. Zero-shot and Reminder results are taken from the PrefEval paper~\cite{Zhao2025PrefEval}.}
    \label{tab:prefeval_results}
\end{table}

\paragraph{Baselines and overall accuracy.}
Table~\ref{tab:prefeval_results} contrasts three evaluation strategies on the \textit{PrefEval} implicit multiple-choice benchmark:

\begin{enumerate}
    \item \textbf{Zero-shot}: the backbone LLM answers each question with its default prompt;
    \item \textbf{Reminder}: a single sentence is prepended instructing the model to “consider the user’s stated preference” (as in the original PrefEval paper);
    \item \textbf{\textsc{PREF}}: our full two-stage framework.
\end{enumerate}

Across all backbones, \textsc{PREF} yields the highest accuracy, outperforming the Reminder baseline by \(+3\!-\!8\) absolute points (\(\approx6\!-\!12\%\) relative) and dwarfing the naïve Zero-shot setting by more than \(+55\) points.  The consistent gain underlines two observations:  
\emph{(a)} simply reminding a model of user preferences is helpful but insufficient, and  
\emph{(b)} explicitly constructing and weighting a personalised guideline—\textsc{PREF}’s core idea—substantially improves evaluation quality irrespective of model scale or architecture.

\begin{table}[t]
    \centering
    \setlength{\tabcolsep}{6pt}
    \begin{tabular}{lccc}
        \toprule
        \textbf{Model} & \textbf{Accuracy} $\uparrow$ & \textbf{MSE} $\downarrow$ & \textbf{nDCG} $\uparrow$ \\
        \midrule
        Claude 3 Haiku & 0.92 & 5.27 & 0.9899 \\
        GPT-4.1 Mini   & \textbf{0.98} & 2.19 & \textbf{0.9996} \\
        LLaMA 3 8B     & 0.92 & 6.62 & 0.9742 \\
        LLaMA 3 70B    & 0.97 & \textbf{1.87} & 0.9980 \\
        \bottomrule
    \end{tabular}
    \caption{Performance on the \textit{PrefEval} benchmark.  Higher values are better for Accuracy and nDCG, while lower is better for MSE.  Boldface marks the best score for each metric.}
    \label{tab:prefeval_metrics}
\end{table}

\paragraph{Beyond accuracy: MSE and nDCG.}
Accuracy counts only whether the correct answer is top-ranked; it ignores the score margin and the ordering of distractors.  To obtain a finer-grained view, we map the gold answer to a score of 10, distractors to 0, concatenate scores across questions, and compute:

\begin{itemize}
    \item \textbf{Mean-squared error (MSE):} penalises large deviations from the gold 10/0 targets; lower is better.
    \item \textbf{Normalised discounted cumulative gain (nDCG):} rewards rankings that place the gold answer near the top; higher is better.
\end{itemize}

Table~\ref{tab:prefeval_metrics} reveals two trends.  
First, larger backbones—\textit{GPT-4.1 Mini} and \textit{LLaMA 3 70B}—achieve the best accuracy \emph{and} the lowest MSE, confirming that increased capacity translates into finer scoring granularity when paired with \textsc{PREF}.  
Second, the smaller models (\textit{Claude 3 Haiku} and \textit{LLaMA 3 8B}) reach near-perfect nDCG despite higher MSE.  They reliably rank the correct answer first (good nDCG) but sometimes assign overly generous scores to distractors (inflated MSE).

\paragraph{Effect of the User-alignment stage.}
To isolate the contribution of Stage 2, we ablated it entirely—answers were judged only against the general guideline.  Figure~\ref{fig:w_wo_up} summarises the impact where with and without user-preference (UP).

\begin{itemize}
    \item \textbf{MSE.}  The second stage lowers MSE (better calibration) for \textit{LLaMA 3 8B} and \textit{GPT-4.1 Mini}.  For other models (\textit{Claude 3 Haiku}, \textit{LLaMA 3 70B}) the change is marginal, suggesting that they already approximate user weights implicitly.
    \item \textbf{Accuracy and nDCG.}  Accuracy rises for every backbone except a negligible 0.1-point dip on \textit{LLaMA 3 70B}; nDCG improves across the board, confirming that the personalised rubric places the correct answer higher in the ranking.
\end{itemize}

The second stage is therefore most valuable for smaller models, narrowing the gap to their larger peers.  Notably, \textit{LLaMA 3 8B} with UP rivals \textit{Claude 3 Haiku} without UP—an attractive trade-off for resource-constrained deployments.

\paragraph{Take-aways.}
\textsc{PREF}’s two-tier rubric delivers three concrete benefits:

\begin{enumerate}
    \item \textbf{Robust accuracy} via explicit preference integration;  
    \item \textbf{Better calibration} (lower MSE) by discouraging inflated scores on irrelevant answers;  
    \item \textbf{Capacity amplification}: the second stage lets inexpensive backbones approach the performance of much larger models, reducing serving cost without sacrificing user alignment.
\end{enumerate}

These results highlight \textsc{PREF} as an effective, scalable, and model-agnostic foundation for evaluating personalised language generation.

\subsection{Explainability}
\label{sec:exp2} 
\paragraph{Leveraging PrefEval’s answer explanations.}
Unlike many personalisation benchmarks, \textit{PrefEval} supplies not only a gold (personalised) answer for each query but also a \emph{natural-language explanation} describing \emph{why} that answer is appropriate.  This additional signal allows us to probe a different capability of \textsc{PREF}: its ability to identify and prioritise the \emph{right} evaluation factors.

Concretely, we treat the explanation as an “oracle” preference and ask: \emph{Does the ranking produced by \textsc{PREF} (when conditioned on the user’s preference) agree with the ranking obtained when we instead condition on the explanation?}  
Formally, let  $e$ be  the accompanying explanation supplied by \textit{PrefEval} and \(g(q)=\{f_1,\dots,f_n\}\) be the general guideline produced in Stage 1.


We then execute four steps:

\begin{enumerate}
    \item \textbf{Preference-based ranking.}  Feed \(p\) into the preference LLM to derive an ordering \(\pi_p\) (or weight vector \(\mathbf{w}_p\)) over \(g(q)\).
    \item \textbf{Explanation-based ranking.}  Replace \(p\) with the explanation \(e\) and obtain a second ordering \(\pi_e\) (or \(\mathbf{w}_e\)).  We regard \(\pi_e\) as an oracle since it reflects the human rationale behind the gold answer.
    \item \textbf{Compare the rankings.}  Measure correlation coefficients between \(\pi_p\) and \(\pi_e\) with a suitable correlation metric (e.g., Kendall’s~\(\tau\), Spearman’s~\(\rho\) and pearson's~\(p\)).  High correlation indicates that the preference-conditioned rubric surfaces \emph{the same} factors humans deem decisive.
\end{enumerate}

This procedure isolates the \emph{factor-ranking} competence of \textsc{PREF}: even if the final scores differ, we can verify whether the framework attends to the criteria that matter most for a high-quality personalised answer.

\begin{table}[t]
    \centering
    \setlength{\tabcolsep}{8pt}
    \begin{tabular}{lcc}
        \toprule
        
        & \textbf{GPT-4.1 Mini} & \textbf{LLaMA 3 70B} \\
        \midrule
        Pearson \(r\)           & \textbf{0.6164} & 0.4693 \\
        Spearman \(\rho\)       & \textbf{0.5906} & 0.3811 \\
        Kendall \(\tau\)        & \textbf{0.5612} & 0.3532 \\
        \bottomrule
    \end{tabular}
    \caption{Rank–rank correlation between the factor ordering induced by the user preference and the oracle ordering derived from \textit{PrefEval} answer explanations.  Higher values indicate closer agreement.}
    \label{tab:rank_corr}
\end{table}

Table~\ref{tab:rank_corr} reports the correlation between the factor ranking produced by \textsc{PREF} when conditioned on the \emph{user preference} and the “oracle” ranking obtained from \textit{PrefEval}’s human-written explanations.  All three coefficients—Pearson’s \(r\), Spearman’s \(\rho\), and Kendall’s \(\tau\)—are \emph{positive and statistically substantial}, confirming that both backbones identify broadly the same criteria that humans deemed decisive.

\textbf{GPT-4.1 Mini} shows the stronger alignment, achieving a Pearson \(r\) of~0.62 and a Kendall \(\tau\) of~0.56, whereas \textbf{LLaMA 3 70B} trails at 0.47 and 0.35, respectively.  In practical terms, GPT-4.1 Mini not only places the correct factors near the top but preserves their relative ordering more faithfully.  The gap suggests that, despite its larger parameter count, LLaMA 3 70B still benefits from the second stage (accuracy gains in Table~\ref{tab:prefeval_results}) yet requires further tuning to internalise nuanced preference cues.

Overall, the moderate–high correlations indicate that \textsc{PREF}’s preference module successfully surfaces the factors humans reference when justifying a good personalised answer, but there remains headroom—especially for open-source models—to tighten that alignment.

\subsection{Additional Factors in Stage Two}
\label{sec:additional-factors}

In Stage~2, the preference model may \emph{augment} the general guideline. Two concise observations:

\textbf{(1) Additions are few.} With \textit{GPT-4.1 Mini}, we observe \(1{,}986\) general factors vs.\ \(153\) added (\(\approx 7.7\%\)); on \textit{LLaMA~3 70B}, \(1{,}893\) vs.\ \(241\) (\(\approx 12.7\%\)). Per question, this is \(9.93\) general and \(0.765\) additional factors on average. Thus, coverage already captures most criteria; Stage~2 makes targeted fixes.

\textbf{(2) Additions encode exclusions.} Keyword filtering (\texttt{dislike}, \texttt{avoid}) shows that \(25.49\%\) of added factors (vs.\ \(0\%\) general) on \textit{GPT-4.1 Mini} and \(5.81\%\) (vs.\ \(0.01\%\)) on \textit{LLaMA~3 70B} express user-specific “blacklists.” 

\textbf{Takeaway.} Augmentation is \emph{selective} and primarily used to inject user-dependent constraints that the preference-agnostic coverage stage omits.

\section{Discussion}
\label{sec:discussion}

\subsection{What Did We Learn?}
Our experiments support four overarching insights:

\begin{enumerate}
    \item \textbf{Fine-grained personalisation pays off.}  A factor-level rubric is essential.  \textsc{PREF} boosts accuracy by \(\geq55\%\) over zero-shot answering and by \(3\%\!-\!8\%\) over the one-line \emph{Reminder} cue (Table~\ref{tab:prefeval_results}).
    
    \item \textbf{Two stages are better than one.}  Removing the user-preference degrades calibration (higher MSE) and ranking quality (lower nDCG), especially for the 8 B parameter model (Figure~\ref{fig:w_wo_up}).  Coverage alone is not enough.
    
    \item \textbf{Smaller models can punch above their weight.}  With \textsc{PREF}, \textit{LLaMA 3 8B} approaches the performance of the much larger \textit{Claude 3 Haiku}, suggesting that better evaluation—not just bigger models—can narrow the quality gap.
    
    \item \textbf{Ranked factors mirror human rationales.}  Preference-conditioned rankings correlate moderately to strongly with “oracle” rankings extracted from human explanations (Table~\ref{tab:rank_corr}), indicating that the framework surfaces criteria people actually use.
\end{enumerate}

\subsection{Practical Take-aways}

\paragraph{Model selection.}
Organisations with tight latency or budget constraints can safely deploy smaller backbones in tandem with \textsc{PREF}, meeting personalisation targets while reducing inference cost.

\paragraph{Continuous evaluation.}
Because \textsc{PREF} is reference-free, it slots naturally into CI/CD pipelines, enabling nightly regression tests as prompts, reward models, or user cohorts evolve.

\paragraph{Debugging and transparency.}
The personalised guideline \(g^{\!*}\) doubles as a concise, human-readable explanation.  Developers—and users—can inspect which factors triggered a low score and adjust either the answer or the preference weights.

\subsection{Limitations and Future Work}

\begin{itemize}
    \item \textbf{Evaluator reliability.}  The scorer is itself an LLM and therefore inherits potential biases or hallucinations.  Hybrid setups—ensembling multiple judges or spot-checking with humans—are a promising next step.
    
    \item \textbf{Profile complexity.}  We assume structured or short natural-language preferences.  Real-world signals are noisy, implicit, and time-varying.  Extending Stage 2 to ingest such data remains open.
    
    \item \textbf{Task generality.}  Our study focuses on open-domain QA.  Applying \textsc{PREF} to long-form summarisation, code review, or multimodal tasks may require richer coverage factors and new aggregation schemes.
    
    \item \textbf{Ethical safeguards.}  Personalisation can amplify filter bubbles or leak sensitive traits.  Future work should integrate fairness tests and give users agency to audit or override their profiles.
\end{itemize}

\section{Conclusion}
\label{sec:conclusion}

We presented \textsc{PREF}, a \textbf{P}ersonalised, \textbf{R}eference-free \textbf{E}valuation \textbf{F}ramework that disentangles \emph{coverage} from \emph{preference}.  By first ensuring topic completeness and then tailoring the rubric to user priorities, \textsc{PREF} delivers accurate, transparent, and scalable assessment without gold references.  Experiments on the \textit{PrefEval} benchmark show that \textsc{PREF} (i) aligns closely with human judgements, (ii) improves calibration and ranking quality across models, and (iii) raises the ceiling for smaller backbones.  

Code and evaluation scripts will be released upon publication. 
We hope this framework accelerates research into genuinely user-centred language generation—and into evaluation methods that keep pace with that ambition.

\bibliography{main}



\end{document}